# Exploring the Performance of Perforated Backpropagation through Further Experiments

Rorry Brenner
Perforated AI
rorry@perforatedai.com

Evan Davis
Skim AI
evan@skimai.com

Rushi Chaudhari
Deloitte
0xrushi@gmail.com

Rowan Morse
University of Pittsburgh
rmm271@pitt.edu

Jingyao Chen
Carnegie Mellon
jingyaoc@andrew.cmu.edu

Xirui Liu
Carnegie Mellon
xiruil@andrew.cmu.edu

Zhaoyi You
Carnegie Mellon
zhaoyiy@cs.cmu.edu

Laurent Itti
University of Southern California
itti@usc.edu

**Abstract.** Perforated Backpropagation is a neural network optimization technique based on modern understanding of the computational importance of dendrites within biological neurons. This paper explores further experiments from the original publication, generated from a hackathon held at the Carnegie Mellon Swartz Center in February 2025. Students and local Pittsburgh ML practitioners were brought together to experiment with the Perforated Backpropagation algorithm on the datasets and models which they were using for their projects. Results showed that the system could enhance their projects, with up to 90% model compression without negative impact on accuracy, or up to 16% increased accuracy of their original models.

## 1 Introduction

The first recording of the action potential of a biological neuron was performed nearly a century ago (Hodgkin, 1939). Once this was discovered, it only took 4 years for a mathematical model to be developed (McCulloch, 1943), which would be the foundation of the perceptron (Rosenblatt, 1958) once computers were capable of processing these ideas. Since then, neuroscience and computer science research have continued to advance, yet the perceptron as the core building block of neural networks has remained the same. Perforated Backpropagation (Brenner, 2025) brings an update to this neuron with the addition of artificial dendrites. The technique is inspired by recent research in neuroscience emphasizing the importance of dendrites (Major, et al., 2013), with some papers even arguing that dendrites, and not neurons, are the fundamental functional unit of the brain (Branco, 2010).

This paper discusses continued experimentation with Perforated Backpropagation. Experiments described here focus the priority on working with models and datasets more researchers in the academic ML community are familiar with, and more researchers in the industry ML community are using for production application. We hope this paper will add additional evidence to the capacity of the technique and encourage more researchers to use it for experiments with the PyTorch systems they are working with. Due to the nature of the hackathon, results will not contain error bars representing repeated runs of the same experiments.

## 2 Background

The human brain's remarkable computational abilities, particularly in the visual system, have long inspired artificial neural networks (ANNs). However, conventional artificial neurons are simplified compared to their biological counterparts, notably lacking the complex processing capabilities provided by dendrites

(Krizhevsky, et al., 2012; Szegedy, et al., 2015; Ciregan, et al., 2012; Kubilius, et al., 2016). In biological neurons, dendrites not only serve as the primary site for synaptic input collection but also perform intricate computations, including passive voltage summation and active spike generation, such as NMDA receptor-mediated spikes (Major, et al., 2013). These dendritic spikes are dynamic, threshold-dependent, and enable single neurons to detect sophisticated patterns that would otherwise require multiple artificial neurons across several layers. This complexity allows biological neurons to encode features that individual neurons within current ANNs cannot learn to code for.

Modern artificial neural networks typically establish direct synaptic connections to neuron cell bodies, using connection weights as their sole parameter, and omit dendritic structures entirely (Widrow & Lehr, 1990). This omission limits their computational power and their ability to match the performance of biological systems, especially in tasks like generalized object recognition. Perforated Backpropagation (PB) introduces additional nonlinear processing layers, serving as artificial dendrites, between each neuron and its presynaptic inputs. These layers are designed to be compatible with PyTorch while remaining distinct from standard backpropagation pathways, thus enabling a new paradigm for neural network learning.

These artificial dendrite nodes behave differently from standard neuron nodes, as the work draws inspiration from the Cascade Correlation algorithm (Fahlman & Lebiere, 1989). Cascade Correlation improves network performance by incrementally adding new hidden nodes that learn to maximize their correlation with the network's error, rather than simply minimizing output error directly. These candidate nodes are trained while the weights in the existing network are frozen, capturing a "snapshot" of the error landscape, and are then integrated as permanent features. This process creates a network architecture where nodes are empowered to perform better at coding for the same feature, a principle that aligns with the goal of introducing dendrite-like modules to enhance the capabilities of artificial neurons. Adding Cascade Correlation, originally designed for use with single layer networks, to deeper models can be compared to the improvements granted by reducing residual error with ResNets (He, et al., 2016), or combining multiple weak learners with boosting (Schapire, 1990). Additional details about this background and related works can be found in our original paper (Brenner, 2025).

# 3 Results

We implement PB in PyTorch, so that it is compatible with many existing models at minimal development cost and effort. In a nutshell, training a PB-enhanced network proceeds as follows: 1) Train the original network until convergence; 2) Freeze the original network's weights and add new PB nodes, then train these with the objective of correcting any remaining errors still made by the frozen original network; 3) Freeze the PB weights, un-freeze the original network's weights, and iterate back to 1) until no further performance improvement is obtained. A more detailed workflow, and details on the loss formulations, are in (Brenner, 2025).

## 3.1 Language Modeling with BERT

The Bidirectional Encoder Representations from Transformers (BERT) model architecture (Kenton, 2019) is a deep learning language model that can be trained or fine-tuned for many tasks. Even in the age of Large Language Models (LLMs), BERT is still a powerful tool that is competitive in efficiency and performance. The architecture offers many versions of different widths and depths, and three different families of models were explored by our research. The first alternative uses robustly optimized pretraining (Liu, 2019). The second uses a Deep Summing Network (DSN) approach. The DSN architecture replaces the computationally expensive self-attention mechanism with simple sum pooling, resulting in 15x faster inference speeds. Without Perforated Backpropagation, DSNs would typically be limited to applications where speed was more critical than accuracy, but with Perforated Backpropagation, these networks become viable for a much broader range of edge and resource-constrained applications where both performance and efficiency matter.

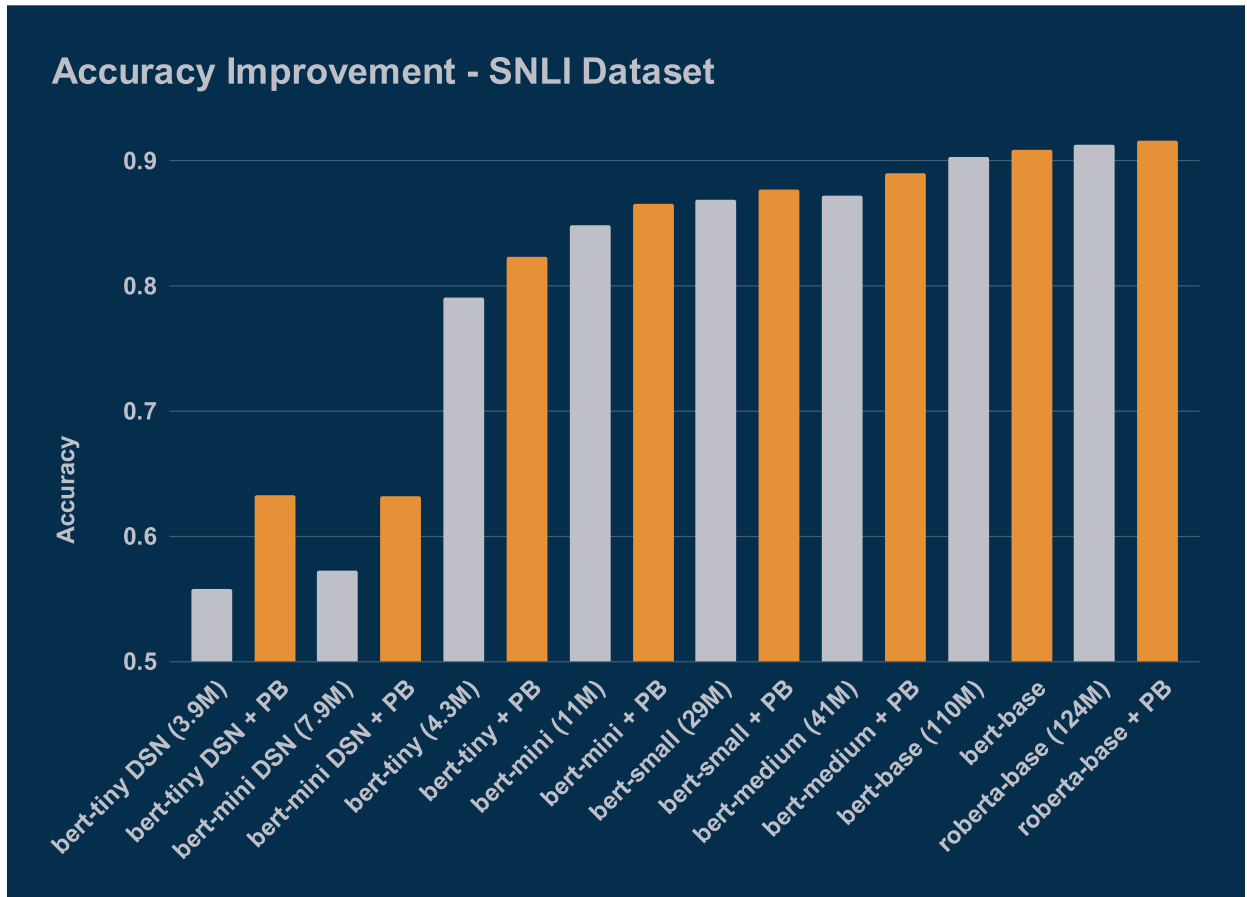

Figure 1. Test scores across all model architectures experimented with on the SNLI dataset. Numbers represent parameter counts. "+ PB" bars represent dendrite optimized models.

Experiments were performed with the SNLI dataset (Bowman, 2015) and the IMDB dataset (Maas, 2011). SNLI consists of 570K samples representing entailment pairs classified into three inference relation types. Experiments with this dataset demonstrated that Perforated Backpropagation improved test set accuracy for BERT variants across a range of model sizes, from 3.9M parameter BERT-tiny DSN to 124M RoBERTa-base. IMDB contains 50K movie reviews with binary sentiment classification labels, and only 25K samples for training. Due to the small dataset size, the BERT models tend to quickly overfit, and so the addition of PB did not benefit larger BERT variants. However, this was an ideal dataset to demonstrate the compression capabilities of PB. A BERT-tiny DSN with dendrites had layers reduced in width by 87.5% (88.7% fewer parameters) and still retained the same accuracy as the original full-width BERT-tiny.

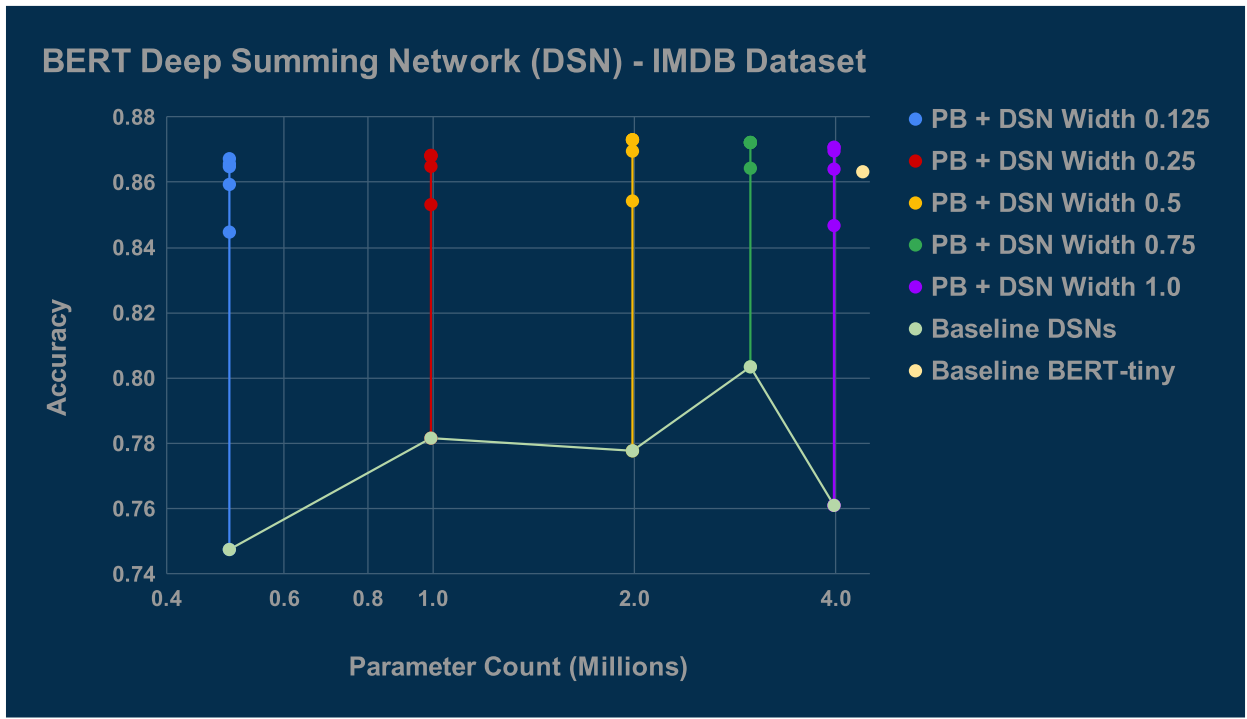

Figure 2. Graph representing compression experiments with DSN compared to BERT-tiny. Baseline BERT-tiny is represented by a single dot, while the baseline DSN's of different widths without dendrites are connected with the bottom line. In each Perforated Backpropagation width setting, additional dots represent accuracy increases as dendrites are added. Parameter increases are minimal in this case because most of the parameters of the DSN model are in the embedding layer, which did not have Dendrites added.

### 3.1.1 PB + DSN vs BERT-tiny Deployment Implications

Fewer Parameters with equal accuracy have two real-world effects when deploying a model for production use-cases. Smaller models can run inference with less hardware and run more effectively on the same hardware. To determine the exact impact of this optimization experiment, the BERT-tiny and PB+DSN Width 0.125 models were run on Google Cloud Platform instances. To simulate the minimal hardware available for edge devices, the first experiment was performed on the smallest available instance, a c2-standard-4 instance that provides only 4 vCPUs and no GPU support. This provided numbers for the total tokens/s throughput that could be processed on restricted hardware. Experiments were then run on scaled up instances to determine the optimal instance type for generating the maximum tokens / dollar. For both models, it turned out the optimal cost instance was a n1-standard-2 with a single T4 GPU.

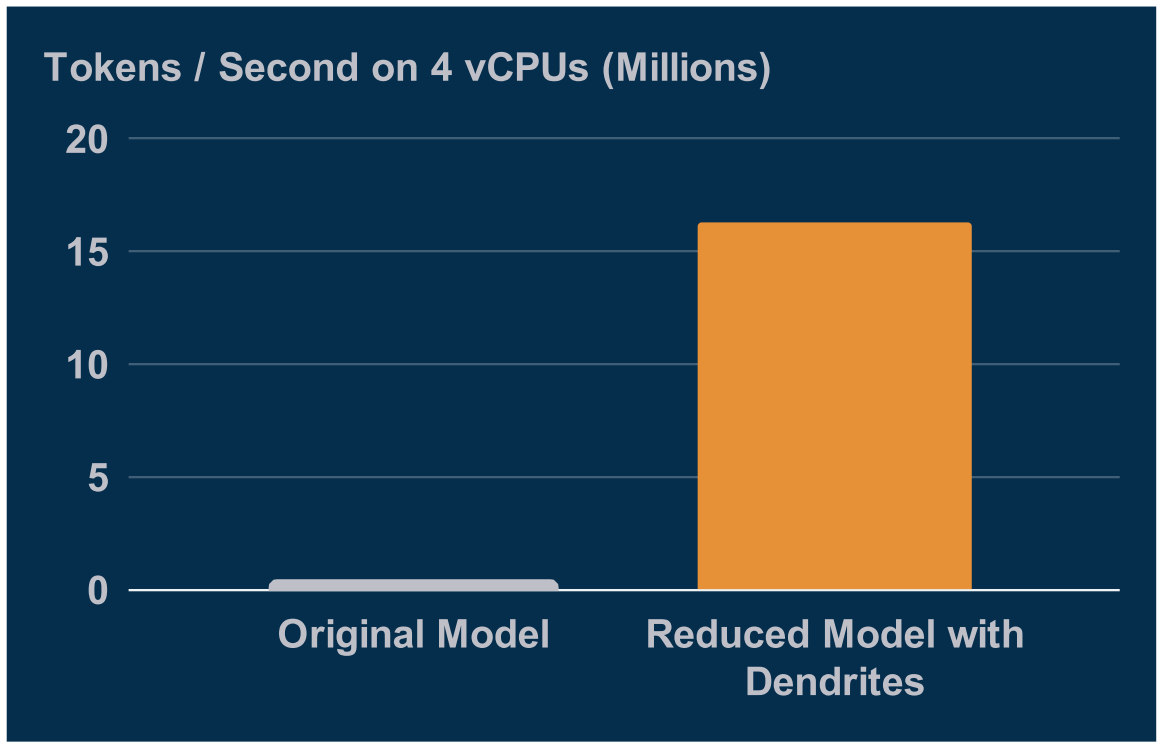

Figure 3. Graph showing tokens / second on Google Cloud c2-standard-4 instance. The order of magnitude of this graph is larger than standard LLMs since BERT is a much smaller model, and therefore the baseline is much faster as well.

Results of this experiment showed that, on restricted CPU-only hardware, the reduced model can process tokens at 158x the speed of the original, 16,319,841 tokens per second vs 107,001. With a T4 GPU the original model will cost 38x more than the compressed model, $0.054 per billion tokens vs $0.0014. It is also worth noting that, on the optimal hardware, the original model can only compute 1,581,885 tokens per second. This means that there are many throughput requirements for which not having access to Perforated Backpropagation would be especially detrimental. In the worst case, a production application might require 16,000,000 tokens per second. In this scenario the original model would require 11 cloud GPUs to perform the same task as a single edge CPU with Perforated Backpropagation.

| VM Type | Hourly Cost | Experiment | Total Parameters | Tokens per second | Cost per B input tokens | Optimal Batch Size |
|---|---|---|---|---|---|---|
| n1-standard-2 (T4 GPU) | $0.31 | Original Model | 4.38M | 1,581,885 | $0.0544 | 3072 |
| | | Reduced Model with Dendrites | 496K | 59,604,227 | $0.0014 | 86016 |
| c2-standard-4 (CPU) | $0.17 | Original Model | 4.38M | 107,001 | $0.4413 | 32 |
| | | Reduced Model with Dendrites | 496K | 16,319,841 | $0.0028 | 768 |

Table 1. Exact values for experiments run on minimal Google Cloud instance, and instance with optimal cost / billion input tokens.

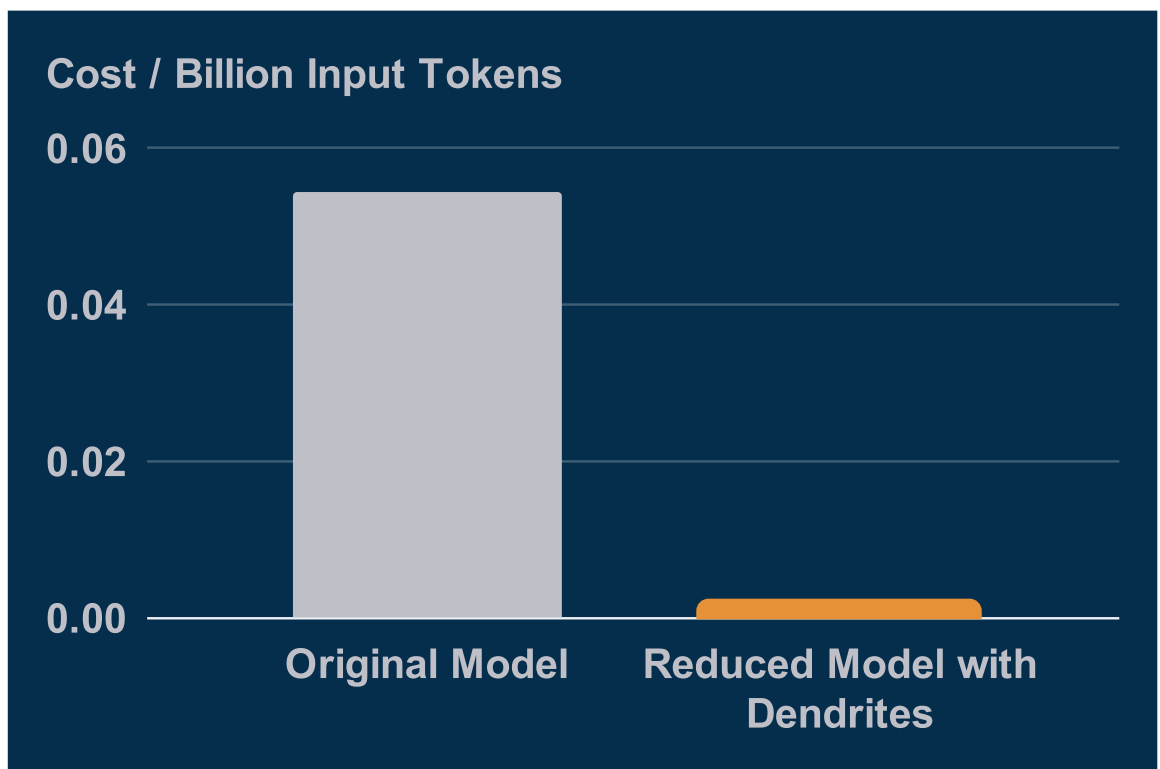

Figure 4. The cost of each token processed on Google Cloud n1-standard-2 instance. Total tokens per second processed were 1,581,885 and 59,604,227 with a cost of $0.31 dollars per hour.

### 3.2 Amino Acid Classification with ProteinBERT

In addition to language modeling, BERT based architectures can be used for amino acid sequence classification. ProteinBERT (Brandes, 2022) served as the foundational model evaluated on the dataset introduced by AMP-BERT (Lee, 2023), which contains 1,778 antimicrobial peptides (AMPs) and an equal number of non-antimicrobial peptides. Each sequence is labelled as either AMP or non-AMP, and the objective is to train a binary classifier that predicts the antimicrobial status of a given protein sequence. Owing to the nature of protein data, the model operates on a compact vocabulary (the 20 canonical amino acids plus special tokens) while accommodating substantially longer sequence contexts.

      This experiment focused on the compression aspect of Perforated Backpropagation's benefits. The two models compared are the original model without optimization and a model reduced in depth from 30 layers to just 12 layers, while simultaneously decreasing the hidden layer width from 1024 to 480. By starting with the reduced model and then adding Dendrites, a new model was able to be created which was comparable in accuracy with only 21% the total parameters as the original.

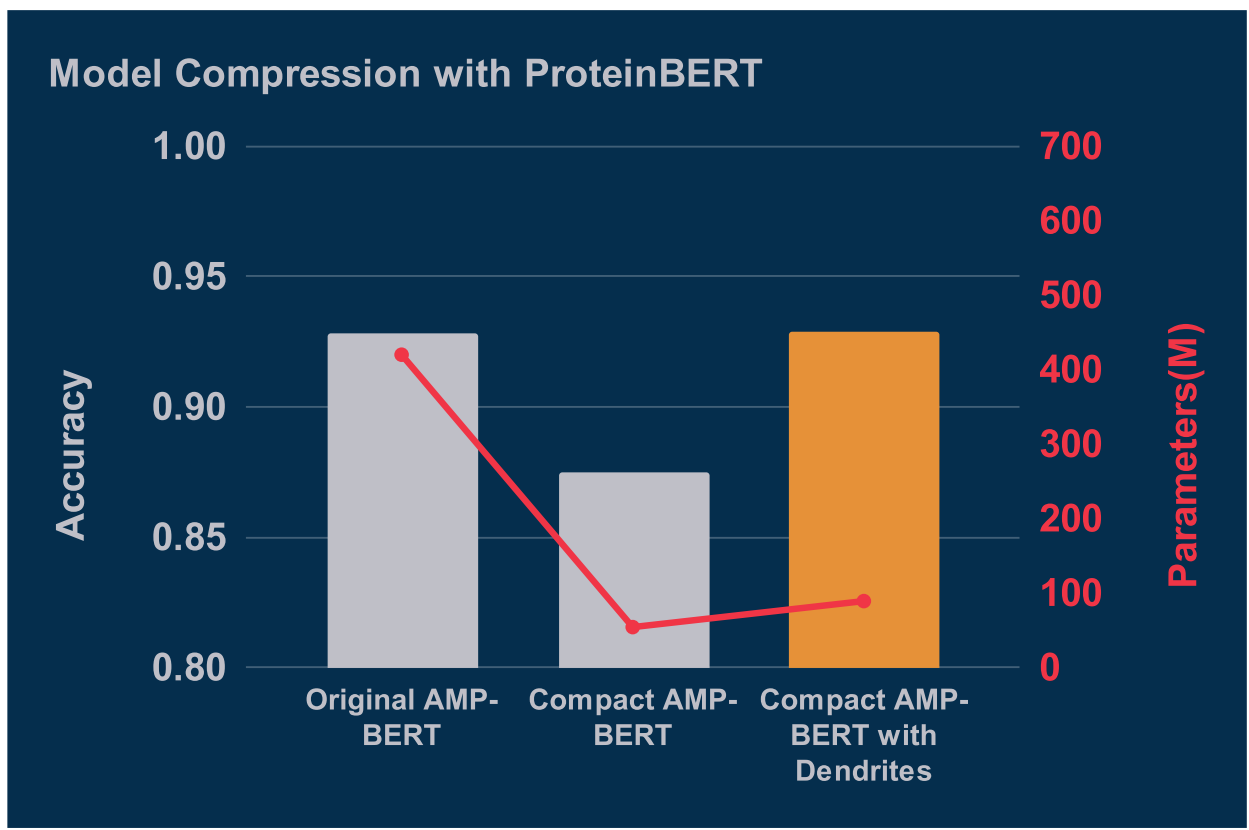

Figure 5. Graph showing the original AMP-BERT model, the compressed version, and the compressed version after the addition of dendrites. Results show that the compact model with dendrites contains a best of both worlds result, with 79% fewer parameters, and equivalent accuracy compared to the original

### 3.3 Further Compression of MobileNet V3

MobileNet (Andrew G. Howard, 2018), now on its third version (Howard, 2019), is a convolutional neural network optimized for resource and accuracy tradeoffs for use in edge and mobile applications. MobileNet is a popular model in the PyTorch model library to use for transfer learning and pretrained backbone components of other architectures. In this study MobileNet V3 was selected to investigate whether a model already optimized for compactness could be further compressed using the Perforated Backpropagation technique. The dataset chosen for this experiment was CIFAR 10 (Krizhevsky, 2009). Experiments were performed both to increase accuracy of the base model and maintain accuracy of a reduced model. Starting with MobileNet-small, a 6% error reduction was achieved by adding Dendrites to the model. By reducing the width of the network by 50% and then adding dendrites, a new architecture was obtained that slightly improved upon the original accuracy while using 35% fewer parameters, reducing the total from 2.54 million to just 1.66.

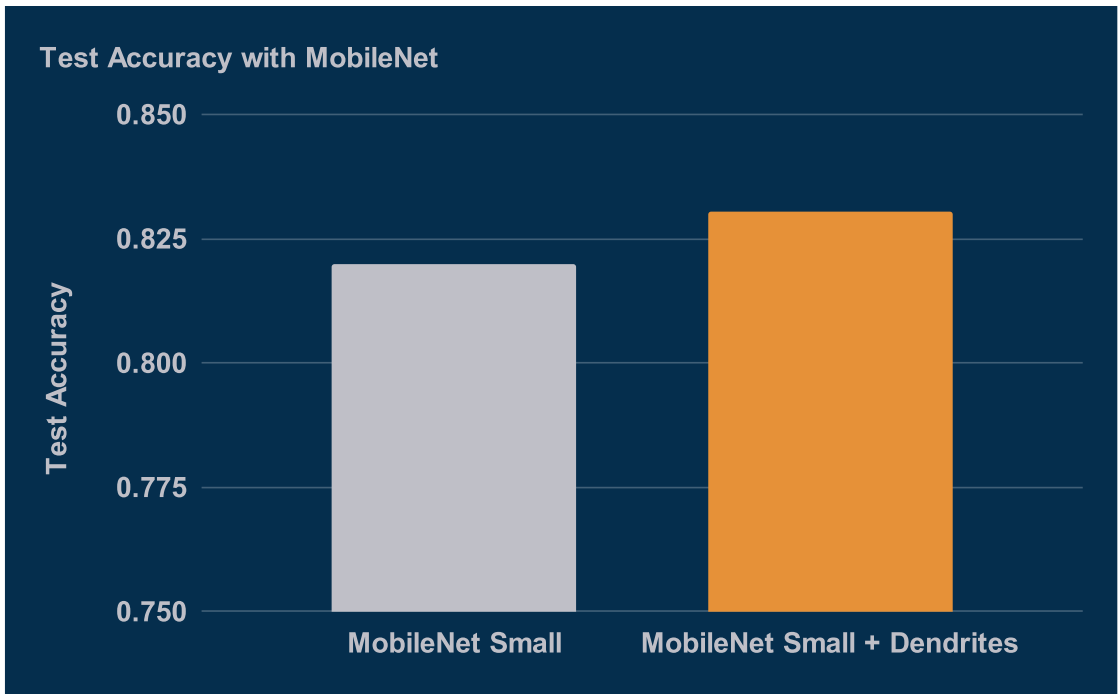

Figure 6. By adding Dendrites to the Small version of MobileNet-V3 accuracy increases from 81.99% to 83.05%.

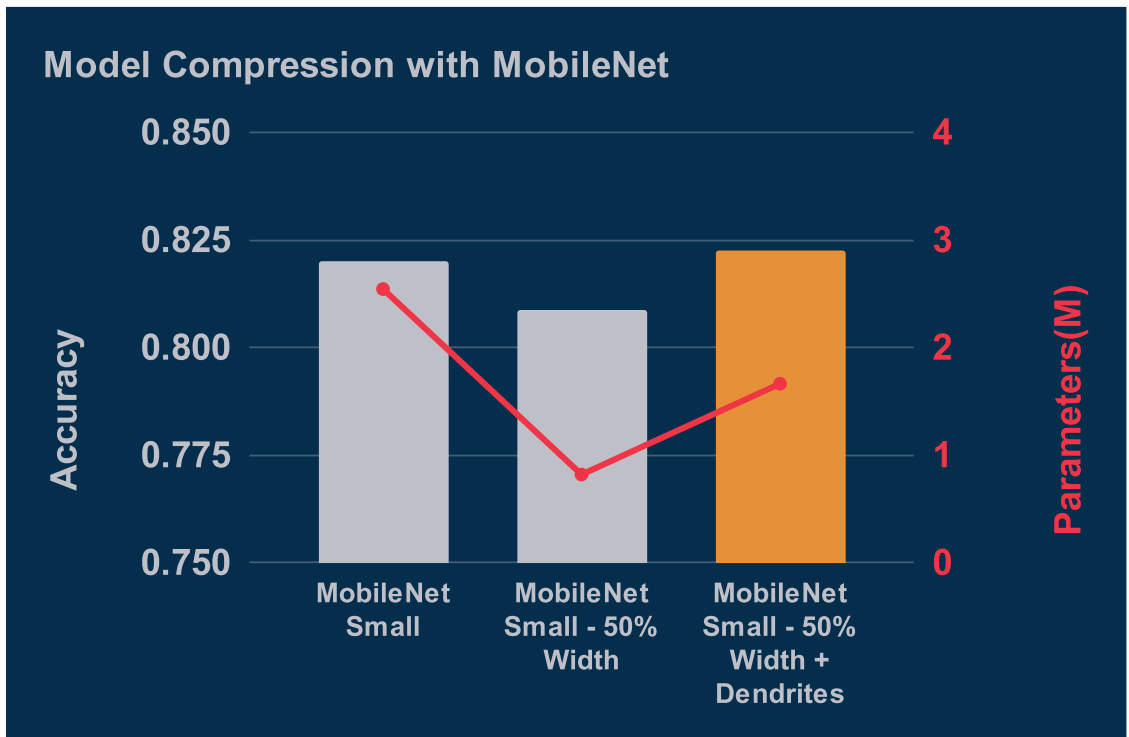

Figure 7. Reducing the width of MobileNet by 50% brings the parameters from 2.54 million to 0.82 million but also reduces the accuracy from 81.99% to 80.8%. By adding Dendrites to this reduced model the accuracy increases slightly higher than the original to 82.25% while the parameter count remains 35% smaller at 1.66 million.

# 4 Conclusion

This work demonstrates that Perforated Backpropagation, inspired by the computational power of biological dendrites, provides a practical and effective means to enhance the efficiency and performance of artificial neural networks. By introducing artificial dendrites as nonlinear processing layers between neuron layers, we show that it is possible to significantly reduce model size while maintaining-or even improving-accuracy across a variety of architectures and domains, including language modeling with BERT, protein sequence classification with ProteinBERT, and image recognition with MobileNet-V3. Our experiments reveal several key benefits:

- **Model Compression with Minimal Accuracy Loss:** Across all tested domains, adding dendrite-inspired modules enabled substantial reductions in parameter count while preserving or improving predictive accuracy.
- **Deployment Efficiency:** The compressed models demonstrated dramatic improvements in inference speed and cost-effectiveness, particularly on resource-constrained hardware.
- **Generalizability:** The approach proved robust across diverse tasks and datasets highlighting its versatility and potential for widespread adoption in both academic and industrial machine learning applications.

These results suggest that Perforated Backpropagation offers a promising path forward for building more biologically inspired and computationally efficient neural networks. By bridging the gap between the simplicity of artificial neurons and the rich processing capabilities of their biological counterparts, this technique enables the design of models that are both lightweight and powerful. Plans for our future experiments include repeating the MobileNet and BERT experiments on their original ImageNet (Deng, et al., 2009) and Wikipedia (Liu, 2018) plus Books Corpus (Zhu, 2015) datasets. We encourage further exploration and adoption of this approach within the PyTorch ecosystem and beyond, with the hope that it will inspire continued innovation at the intersection of neuroscience and machine learning.

## 4.1 Reproducing Results and Using our System

Instructions to reproduce the results of this paper can be found in our public GitHub repository at https://github.com/PerforatedAI/PerforatedAI-Examples. Instructions and best practices to add Perforated Backpropagation to any project built with PyTorch can be found in our second GitHub repository at https://github.com/PerforatedAI/PerforatedAI-API.

# References

Andrew G. Howard, M. Z. B. C. D. K. W. W. T. W. M. A. H. A., 2018. Mobilenets: Efficient convolutional neural networks for mobile vision applications.. *arXiv,* Volume 1704.04861.

Bowman, S. R. a. A. G. a. P. C. a. M. C. D., 2015. A large annotated corpus for learning natural language inference. *Proceedings of the 2015 Conference on Empirical Methods in Natural Language Processing (EMNLP).*

Branco, T. a. M. H., 2010. The single dendritic branch as a fundamental functional unit in the nervous system. *Current opinion in neurobiology ,* Volume 20.4, pp. 494-502.

Brandes, N. e. a., 2022. ProteinBERT: a universal deep-learning model of protein sequence and function.. *Bioinformatics,* Volume 30.8, pp. 2102-2110.

Brenner, R. a. L. I., 2025. "Perforated Backpropagation: A Neuroscience Inspired Extension to Artificial Neural Networks.. *arXiv preprint arXiv:2501.18018.*

Ciregan, D., Meier, U. & Schmidhuber, J., 2012. *Multi-column deep neural networks for image classification.* s.l., s.n., p. 3642–3649.

Deng, J. et al., 2009. *Imagenet: A large-scale hierarchical image database.* s.l., s.n., p. 248–255.

Fahlman, S. & Lebiere, C., 1989. The cascade-correlation learning architecture. *Advances in neural information processing systems,* Volume 2.

He, K., Zhang, X., Ren, S. & Sun, J., 2016. *Deep residual learning for image recognition.* s.l., s.n., p. 770–778.

Hodgkin, A. L. a. A. F. H., 1939. Action potentials recorded from inside a nerve fibre.. *Nature,* Volume 144.3651, pp. 710-711.

Howard, A. e. a., 2019. Searching for mobilenetv3. *Proceedings of the IEEE/CVF international conference on computer vision.*

Kenton, J. D. M.-W. C. a. L. K. T., 2019. Bert: Pre-training of deep bidirectional transformers for language understanding.. *Proceedings of naacL-HLT,* 1(2).

Krizhevsky, A. a. G. H., 2009. Learning multiple layers of features from tiny images..

Krizhevsky, A., Sutskever, I. & Hinton, G. E., 2012. Imagenet classification with deep convolutional neural networks. *Advances in neural information processing systems,* Volume 25.

Kubilius, J., Bracci, S. & Op de Beeck, H. P., 2016. Deep neural networks as a computational model for human shape sensitivity. *PLoS computational biology,* Volume 12, p. e1004896.

Lee, H. e. a., 2023. AMP-BERT: Prediction of antimicrobial peptide function based on a BERT model. *Protein Science,* 32.1(e4529.).

Liu, P. J. e. a., 2018. Generating wikipedia by summarizing long sequences.. *arXiv preprint arXiv:1801.10198.*

Liu, Y. e. a., 2019. Roberta: A robustly optimized bert pretraining approach.. *arXiv preprint arXiv:1907.11692.*

Maas, A. L. a. D. R. E. a. P. P. T. a. H. D. a. N. A. Y. a. P. C., 2011. Learning Word Vectors for Sentiment Analysis. *Proceedings of the 49th Annual Meeting of the Association for Computational Linguistics: Human Language Technologies,* pp. 142-150.

Major, G., Larkum, M. E. & Schiller, J., 2013. Active properties of neocortical pyramidal neuron dendrites. *Annual review of neuroscience,* Volume 36, p. 1–24.

McCulloch, W. S. a. W. P., 1943. A logical calculus of the ideas immanent in nervous activity.. *The bulletin of mathematical biophysics,* Volume 5, pp. 115-133..

Rosenblatt, F., 1958. *The perceptron: a probabilistic model for information storage and organization in the brain..* s.l.:Psychological review 65.6 386..

Schapire, R. E., 1990. The strength of weak learnability. *Machine learning,* Volume 5, pp. 197-227.

Szegedy, C. et al., 2015. *Going deeper with convolutions.* s.l., s.n., p. 1–9.

Widrow, B. & Lehr, M. A., 1990. 30 years of adaptive neural networks: perceptron, madaline, and backpropagation. *Proceedings of the IEEE,* Volume 78, p. 1415–1442.

Zhu, Y. e. a., 2015. Aligning books and movies: Towards story-like visual explanations by watching movies and reading books. *Proceedings of the IEEE international conference on computer vision.*